%% file: samplepaper.tex
\documentclass[runningheads]{llncs}
\usepackage[T1]{fontenc}
\usepackage{graphicx}
\usepackage{cite}
\usepackage{textcomp}
\usepackage{xcolor}
\usepackage{amsfonts, amsmath, amssymb}
\usepackage{algorithm}
\usepackage{algorithmic}
\usepackage{multirow}
\usepackage{tabularx}
\usepackage{booktabs}
\usepackage{float}
\usepackage{color, colortbl}
\usepackage[pagebackref,breaklinks,colorlinks]{hyperref}
\usepackage{url}
\usepackage{color}

\begin{document}
\title{Adaptive Low Rank Adaptation of Segment Anything to Salient Object Detection}
%
%
\author{Ruikai Cui \and
Siyuan He \and
Shi Qiu
\authorrunning{R. Cui \emph{et al.}}
%
\institute{Australian National University\\
\email{\{ruikai.cui, siyuan.he, shi.qiu\}@anu.edu.au}}
}
\maketitle              
\begin{abstract}

Foundation models, such as OpenAI's GPT-3 and GPT-4, Meta's LLaMA, and Google's PaLM2, have revolutionized the field of artificial intelligence. A notable paradigm shift has been the advent of the Segment Anything Model (SAM), which has exhibited a remarkable capability to segment real-world objects, trained on 1 billion masks and 11 million images. 
Although SAM excels in general object segmentation, it lacks the intrinsic ability to detect salient objects, resulting in suboptimal performance in this domain. To address this challenge, we present the Segment Salient Object Model (SSOM), an innovative approach that adaptively fine-tunes SAM for salient object detection by harnessing the low-rank structure inherent in deep learning. Comprehensive qualitative and quantitative evaluations across five challenging RGB benchmark datasets demonstrate the superior performance of our approach, surpassing state-of-the-art methods.

\end{abstract}
\section{Introduction} 
\label{sec:intro}
Foundation models~\cite{gpt3_brown,llama_touvron,sam_kirillov} have received significant interests in recent years, owing to their exceptional performance across a multitude of diverse tasks 
These models typically consume billions of parameters, trained on expansive web-scaled datasets for fundamental tasks such as next token prediction~\cite{bert_devlin} or masked region completion~\cite{mae_he}. A particularly compelling instance of these models is the Segment-Anything Model (SAM)~\cite{sam_kirillov}, which has been trained on an unprecedentedly vast dataset comprising 11 million images and 1 billion masks.

Despite the Segment-Anything Model's (SAM) noteworthy proficiency in generating masks to segment real-world objects, it is deficient in the detection of salient objects. 
This shortcoming leads to suboptimal performance in isolating a single salient object from a given RGB image, a crucial aspect of computer vision that emphasizes the identification of the most visually striking or attention-demanding object within an image. 

Traditional approaches for harnessing the capabilities of foundation models for downstream tasks generally include fine-tuning the entire model~\cite{finetuning_howard} or integrating additional adapter layers~\cite{pe_houlsby}. However, most foundation models possess dimensions that render them unfeasible for fine-tuning on a typical consumer-grade hardware. 
For instance, the fine-tuning of LLaMA~\cite{llama_touvron}, a model with a staggering 65 billion parameters, requires more than 780G of GPU memory, equivalent to ten A100 80G GPUs. Such computational demands are prohibitive for most organizations or individuals.

To this end, we introduce the Segment Salient Object Model (SSOM), an adaptive fine-tuning of the Segment-Anything Model (SAM) tailored to the salient object detection problem. Specifically, we leverage the low-rank structure~\cite{lora_hu} inherent in deep learning models and employ the Adaptive Low Rank Adaptation (AdaLoRA) strategy~\cite{adalora_zhang} to fine-tune SAM for salient object detection tasks. The AdaLoRA strategy capitalizes on the low-rank characteristics of deep neural networks when applied to downstream tasks and adaptively assigns rank importance to the learnable weight matrices. This technique reduces the number of learnable parameters from 93M to 4M, allowing us to effectively fine-tune SAM on salient object detection datasets, thereby empowering SAM to identify saliency in RGB images. 

The principal contributions of our study are threefold. Firstly, to the best of our knowledge, our work pioneers the application of foundation models for the task of salient object detection. Secondly, we present the novel use of adaptive low rank adaptation to address the complexities inherent in fine-tuning the Segment-Anything Model. Lastly, we undertake comprehensive experiments across various benchmark datasets to verify the efficacy and efficiency of our proposed methodology.

The remaining content of this paper is structured as follows: Sec.~\ref{sec:related_work} offers a review of related literature. In Sec.~\ref{sec:method}, we explain the details of our proposed methodology. Sec.~\ref{sec:exp} presents our experimental approach and discusses the chosen evaluation metrics. Finally, Sec.~\ref{sec:conclusion} concludes the paper with a discussion on the limitations of our method.

\section{Related Works}
\label{sec:related_work}
\subsection{Salient Object Detection}
Salient Object Detection (SOD)~\cite{salod_zhou,survey_borji} focuses on the identification and segmentation of the most visually prominent objects within RGB or RGBD images. Traditional SOD methodologies involved the extraction of low-level features coupled with the integration of multi-modal information~\cite{search_ltti}. However, the advent of deep learning techniques~\cite{resnet_he,attn_vaswani} has considerably reshaped the SOD landscape, leading to the predominance of supervised methods. Such supervised techniques~\cite{edn_wu,pfsnet_ma,gatenet_zhao} for SOD exploit large annotated datasets to train deep learning models, facilitating the capture and comprehension of complex visual patterns intrinsic to salient object detection. Nevertheless, these methods inherently depend on the availability and quality of labeled training data. To mitigate potential issues of data scarcity, unsupervised learning methods~\cite{usod_zhang,deepups_nguyen,weak_zhang} for SOD were introduced. These techniques typically focus on enhancing the coarse saliency cues extracted by traditional SOD methods with low-level cues, thereby reducing the necessity for an extensive labeling process inherent in data collection for supervised methods. Contrary to these conventional approaches, our work introduces a novel paradigm for SOD that harnesses the power of large-scale pre-trained models. By leveraging the superior feature extraction and generalization capabilities of such models, we aim to enhance the performance of SOD via the fine-tuning of those foundation models.

\subsection{Foundation Models}

Foundation Models~\cite{foundation_zhou} are a class of deep learning models that have been pre-trained on vast amounts of data, acquiring broad capabilities and knowledge that can be utilized in diverse downstream applications through fine-tuning processes. Starting from Natural Language Processing (NLP), foundation models in NLP has proven effective in enriching language models with sophisticated knowledge beneficial for downstream tasks, such as the understanding of long-term dependencies and hierarchical relationships. A prime example is ChatGPT\footnote{https://openai.com/blog/chatgpt}, fine-tuned from the generative pretrained transformer GPT-3.5~\cite{gpt3_brown}, which demonstrated superior performance when applied reinforcement learning from human feedback \cite{rlhf_christiano} and is seen as a significant step towards artificial general intelligence. Stimulated by these advancements in NLP, CLIP~\cite{clip_radford} employs contrastive learning to train text and image encoders to align these two modalities. Upon training, CLIP, as a foundation model, facilitates zero-shot generalization to novel visual concepts and data distributions~\cite{dalle_ramesh}. The Segment Anything Model (SAM)~\cite{sam_kirillov} is another exemplar of foundation models, recently proposed as a potent tool for image segmentation. 

\subsection{Parameter-Efficient Adaptation}

The impressive generalization capabilities of foundation models foster the allure of applying transfer learning to adapt these models for specific downstream tasks of interest. However, the escalating computational demands and the size of modern deep learning models~\cite{gpt3_brown,llama_touvron,clip_radford} challenge traditional transfer learning strategies~\cite{transfer_zhuang}, leading to a burgeoning demand for parameter-efficient adaptations. One approach entails the insertion of adapter layers~\cite{adapter_houlsby,norm_lin} within the existing layers of a neural network, as these layers can be trained independently from the base model, thus reducing the parameter requirement for fine-tuning. Despite their adaptability, these supplementary adapter layers often contribute to latency during inference. In contrast to these methodologies, LoRA~\cite{lora_hu} investigates the low-rank structure~\cite{rank_li} inherent in deep learning and proposes to adapt foundation models by exclusively updating low-rank parameterized weight matrices. Subsequently, AdaLoRA~\cite{adalora_zhang} identified that earlier low-rank fine-tuning methods~\cite{lora_hu,qlora_dettmers} often distributed the weight update budget uniformly across all pre-trained weight matrices, neglecting the varying importance of different weight parameters. Consequently, they proposed the adaptive allocation of the parameter budget among weight matrices in accordance with their respective importance scores. 

\section{Segment Salient Object Model}
\label{sec:method}

\subsection{Overview}

Given an RGB image, denoted as $ X \in \mathbb{R}^{H\times W \times 3}$, a salient object is defined as the object that most prominently stands out to an observer. Formally, a salient object is represented as a binary segmentation mask $Y\in \mathbb{R}^{H\times W}$, in which a value of $1$ denotes the salient region and a value of $0$ demarcates the non-salient regions.

Our goal is developing an effective method for extracting the salient object from a given image. To this end, we propose a novel approach, the Salient Segment-Only Model (SSOM), designed to specifically extract salient objects from RGB images. The SSOM is an innovative model designed specifically for the extraction of salient objects from RGB images. 
The SSOM builds upon the fundamental architecture of SAM, composed of four principal components: the image encoder $E_i$, the prompt encoder $E_p$, the mask decoder $D_m$, and Adaptive Low Rank Adaptation (AdaLoRA) layers $l_{ada}$.


Taking advantage of the inherent capability of the foundation model SAM to manage tasks within the general domain, we introduce AdaLoRA layers into the image encoder to specialize its functionality towards saliency detection. We employ the pre-trained weights of SAM for initialization and selectively freeze the image encoder weights. The training process is confined to the AdaLoRA layers, the mask decoder, and the prompt encoder, leading to an efficient adaptation of SAM to saliency detection.

In the ensuing subsections, we provide a detailed description of each component's role and functionality within the proposed architecture.

\subsection{SAM for Saliency}

In this section, we elaborate on our adaption of the Segment Anything Model (SAM) to be specifically tailored for saliency detection.

We preserve the fundamental architecture of SAM, which includes the image encoder, the prompt encoder, and the mask decoder. However, we introduce vital modifications to these components to adapt them for the specialized task of salient object detection. Firstly, we incorporate Adaptive Low Rank Adaptation (AdaLoRA) layers into the image encoder, which we will detail further in Sec.\ref{sec:adalora}. Secondly, we introduce a learnable embedding, utilized as the saliency prompt.

Our proposed Salient Segment-Only Model (SSOM) operates as follows: given an input image $X \in \mathbb{R}^{H\times W \times 3}$, the image is supplied to the image encoder $E_i$ and encoded into $k$ tokens of size $\mathbb{R}^n$, where $n$ represents the embedding dimension. The image encoder performs a \textit{patchification} process that divides an image into $16 \times 16$ patches, each of which is then embedded as a token using a convolution layer. Subsequently, the image encoder applies transformer blocks~\cite{attn_vaswani} $l$ times to produce the final tokens. In contrast to SAM, which accepts user-provided prompts, we introduce a learnable embedding $\mathcal{S}$ of size $\mathbb{R}^n$ as the prompt, serving to indicate the concept of saliency. The mask decoder accepts both the image embedding and the prompt embedding to generate a binary mask $M \in \mathbb{R}^{H\times W}$, which represents the salient object in the image.

To initiate the model, we use SAM pre-trained weights to initialize the image encoder, mask decoder, and prompt encoder, but we deliberately freeze the image encoder to circumvent the computationally expensive training of this heavyweight component. In the subsequent section, we will introduce a parameter-efficient fine-tuning strategy designed to adapt SAM for the task of saliency detection.

\subsection{AdaLoRA parameter-efficient fine-tuning}
\label{sec:adalora}

Our objective is to implement incremental updates to the pre-trained weights of the image encoder in a parameter-efficient manner with AdaLoRA~\cite{adalora_zhang}. In a typical neural network, weight matrices of many dense layers are usually full-rank. When adapted to specific tasks, Aghajanyan~\emph{et al.}\cite{rank_aghajanyan} demonstrated the low 'intrinsic dimension' property for pre-trained language models, a concept that inspired Hu\emph{et al.}~\cite{lora_hu}. They proposed updating a pre-trained weight matrix $W_0 \in \mathbb{R}^{d \times k}$ as $W_0 + \Delta W$, where $\Delta W \in \mathbb{R}^{d \times k}$ is a low-rank matrix decomposed as $\Delta W = BA$. Here, $B\in\mathbb{R^{d\times r}}$, $A \in \mathbb{R}^{r \times k}$ and the rank $r \ll \min(d, k)$. During training, the pre-trained weights remain frozen and $\Delta W$ serves as the trainable parameter. By decomposing it as the product of two low-rank matrices, we can effectively reduce the memory and computational cost of fine-tuning.

Consider the image encoder $E_i$, which contains $l$ transformer blocks. The AdaLoRA-based fine-tuning of each block can be formulated as:
\begin{align}
h = W_{0}x + \Delta Wx = W_{0}x + P\Lambda Qx.
\end{align}
Here, $W_{0} \in \mathbb{R}^{d \times k}$ represents the pre-trained weight matrix of the block, and $x$ denotes the input to this transformer block, with $h$ as the output. $W_{0}$ is the frozen weight, and we update this block with $\Delta W$ parameterized with $P \in \mathbb{R}^{d \times r}$ and $Q \in \mathbb{R}^{r \times k}$. These matrices represent the left/right singular vectors of $\Delta W$, while the diagonal matrix $\Lambda \in \mathbb{R}^{r \times r}$ contains the singular values ${\lambda_i}_{1 \leq i \leq r}$ with $r \ll \min(d, k)$.

More specifically, we employ AdaLoRA to adapt the weight matrices of the multi-head attention mechanism within each transformer block. This restructures the attention process as follows:
\begin{align}
\text{Attention}(Q, K, V) & = \text{softmax}\left(\frac{QK^T}{\sqrt{k}}\right)V
\end{align}
where
\begin{align}
Q & = (W_q + P_q\Lambda_qQ_q)x \\
K & = W_kx \\
V & = (W_v + P_v\Lambda_vQ_v)x.
\end{align}
We maintain the projection weight matrices $W_q, W_k, W_v$, and $A_q$ as frozen, while $P_q, \Lambda_q, Q_q$, and $P_v, \Lambda_v, Q_v$ serve as the trainable AdaLoRA parameters.

By fine-tuning the pre-trained SAM model with the AdaLoRA strategy, we are able to better adapt the model to specific tasks. This approach yields a more effective and efficient model for saliency detection.

\subsection{Loss functions}

In defining our training objective function, we make use of a combination of Binary Cross Entropy (BCE) loss and Intersection over Union (IoU) loss. Let us denote $y$ as the ground truth salient object mask and $\hat{y}$ as the predicted salient object mask. The BCE loss and IoU loss can be mathematically represented as:
\begin{align}
\mathcal{L}_{BCE} & = -\frac{1}{W \times H} \sum_{i=1}^{W \times H} y_i \log \hat{y}_i + (1 - y_i) \log(1 - \hat{y}i) \\
\mathcal{L}_{IoU} & = 1 - \frac{TP}{FN + TP + FP},
\end{align}
where TP, FP, and FN denote true-positive, false-positive, and false-negative results, respectively.

Following the approach proposed by Zhang~\emph{et al.}~\cite{adalora_zhang}, we introduce an orthogonality regularizer to preserve the orthogonality within the singular vector matrices $P$ and $Q$. This regularizer, $R(P, Q)$, guarantees that the transpose of $P$ equals $Q$ and conversely. It can be defined as:
\begin{equation}
R(P, Q) = ||P^{T}P - I||_{F}^{2} + ||QQ^{T} - I||_{F}^{2},
\end{equation}
where $I$ symbolizes the identity matrix, and $||\cdot||_{F}$ stands for the Frobenius norm.

Our total loss function $\mathcal{L}$ can then be defined as a weighted amalgamation of the aforementioned loss components. With equal weighting for the BCE loss and IoU loss, $\mathcal{L}$ can be formulated as:
\begin{equation}
\mathcal{L} = \mathcal{L}_{BCE} + \mathcal{L}_{IoU} + \lambda R(P, Q).
\end{equation}
Here, $\lambda$ functions as a hyperparameter, dictating the influence of the orthogonality regularizer on the total loss. Essentially, it determines the intensity with which we enforce the orthogonality condition on matrices $P$ and $Q$.

\subsection{Training Strategy}

The process of the training strategy is summarized in the following pseudo-code:

\begin{algorithm}
\caption{Training Strategy}
\label{alg:train}
\begin{algorithmic}[1]
\REQUIRE Dataset $D$; total iterations $T$; budget schedule $\{b^{(t)}\}_{t=0}^{T}$;
\FOR{$t = 1, \dots , T$}
    \STATE Sample a mini-batch $\mathcal{X}, \mathcal{Y}$ from $D$ and compute the output $\hat{\mathcal{Y}} = \text{SSOM}(\mathcal{X})$;
    \STATE Compute the gradient $\nabla \mathcal{L}(\mathcal{Y}, \hat{\mathcal{Y}})$;
    \STATE Update learnable parameters in $D_m$ and $E_p$;
    \STATE Update $P^{(t+1)} = P^{(t)} - \eta\nabla_{P} \mathcal{L}$ and $Q^{(t+1)} = Q^{(t)} - \eta\nabla_{Q} \mathcal{L}$;
    \STATE Update $\Lambda^{(t+1)} = \mathcal{T} (\Lambda^{(t)} - \eta\nabla_{\Lambda} \mathcal{L}, S_{k}^{(t)})$ given the budget $b^{(t)}$;
\ENDFOR
\end{algorithmic}
\end{algorithm}

To elucidate further, initially, we load the pre-trained weights from the SAM model. Then, the weights of the image encoder are frozen, and the model is trained only on the $E_p$, $D_m$, and AdaLoRA layers. During the training procedure, as outlined in Algorithm \ref{alg:train}, the model makes predictions on the training data, and the corresponding weights are updated based on the computed gradients.

Following Zhang et al.~\cite{adalora_zhang}, we include a Rank Importance Allocator. At each training step $t$, the singular values are first updated through a stochastic gradient step, as defined by:
\begin{equation}
    \tilde{\Lambda}^{(t)} = \Lambda^{(t)} - \eta \nabla_{\Lambda} \mathcal{L},
\end{equation}
where $\eta > 0$ represents the learning rate.
Subsequently, given the importance score $S^{(t)}$, the singular values undergo pruning according to:
\begin{equation}
    \Lambda^{(t+1)} = \mathcal{T} (\tilde{\Lambda}^{(t)}, S^{(t)}) \text{ with }
    \mathcal{T} (\tilde{\Lambda}^{(t)}, S^{(t)})_{ii} =
\begin{cases}
\tilde{\Lambda}^{(t)}_{ii} & \text{if } S^{(t)}_{i} \text{ in the top-} b^{(t)} \text{ of } S^{(t)} \\
0 & \text{otherwise}
\end{cases}
\end{equation}
Here, $S^{(t)}$ encompasses the importance score of all triplets $(P, \Lambda, Q)$. The quantity $b^{(t)}$ denotes the budget for remaining singular values at step $t$. In this manner, greater budget is allocated to incremental matrices with higher importance, achieved by pruning the singular values of less significant ones. As the most straightforward method to quantify the importance of each triplet, we use the magnitude of singular values, \emph{i.e.}, $S_{i} = |\lambda_{i}|$.

Lastly, the model parameters are updated, and this process is iterated until the model converges or specified iteration limit.

\section{Experiments}
\label{sec:exp}
\subsection{Implementation Details}
We leverage the official pre-trained weights provided for the SAM model in our implementation. The optimization process employs the Adam optimizer. Our model undergoes training for a total of 30 epochs, commencing with an initial learning rate of $10^{-4}$. To cater to the evolving learning.

\subsection{Experimental Setup}
\subsubsection{Datasets}
In line with the established protocols for RGB SOD benchmarking, our experimental framework employs the DUTS-TR~\cite{duts_wang} and MSB-TR~\cite{msra_wang} datasets for training, providing a comprehensive range of images to assure the versatility of our proposed model.
We assess the performance of our method on a diverse selection of datasets, including HKU-IS~\cite{hku_is_li}, PASCAL-S~\cite{pascals_li}, ECSSD~\cite{ecssd_yan}, DUTS-TE~\cite{duts_wang}, DUT-OMRON~\cite{dut_yang}, and MSB-TE~\cite{msra_wang}. For comparative analysis, we include several contemporary methods~\cite{scrn_wu,gatenet_zhao,itsd_zhou,ldf_wei,ctd_zhao,pfsnet_ma,edn_wu} in our study.
The training set is comprised of approximately 27,106 images, and the test set constitutes around 37,568 images. Each dataset offers unique testing scenarios, thereby facilitating a comprehensive evaluation of our model's performance across varied conditions.

\subsubsection{Evaluation Metrics}
In assessing the performance of our model, we resort to two widely accepted evaluation metrics: the F-measure ($F_\beta$) and the Mean Absolute Error ($\mathcal{M}$).
The F-measure, a harmonic mean of precision and recall, is formally expressed as:
\begin{equation}
F_\beta = \frac{(1 + {\beta}^2) \times \text{Precision} \times \text{Recall}}{{\beta}^2 \times \text{Precision} + \text{Recall}}
\end{equation}
Meanwhile, the Mean Absolute Error, a measure of the absolute difference between the ground truth and the predicted saliency map, is given by:
\begin{equation}
\mathcal{M} = \frac{1}{W \times H} \sum_{i=1}^{W} \sum_{j=1}^{H} |G_{i, j} - P_{i, j}|
\end{equation}
where $W$ and $H$ denote the width and height of an image, respectively, and $G$ and $P$ represent the ground truth and predicted saliency map, correspondingly. This combination of metrics provides a robust and comprehensive evaluation of our model's performance.

\subsection{Quantitative result}

We conducted a comparative analysis of our method with seven state-of-the-art RGB Saliency Object Detection (SOD) methods. The performance assessment took into consideration five benchmark datasets and was based on the F-Measure and Mean Absolute Error (MAE) metrics. Detailed results are tabulated in Tab.~\ref{tab:benchmark}.

Our model exhibited superior performance, outperforming comparative methods on most metrics across the evaluated datasets. Remarkably, our model procured the highest scores in both $F_\beta$ and $\mathcal{M}$ metrics on three over five datasets. These outcomes underscore the efficacy of integrating foundation models into saliency detection tasks. Although there is a minor disparity in performance on the PASCAL-S and HKU-IS dataset, our methodology, in a broader perspective, epitomizes the substantial potential of employing this technique for augmenting the precision of saliency detection.

\begin{table*}
    \centering
    \caption{Experimental results on SOD benchmarks. Best scores are in marked in bold.}
    \include{files/benchmark}
    \label{tab:benchmark}
\end{table*}

\section{Conclusion}
\label{sec:conclusion}
In conclusion, our proposed model, integrating a novel fine-tuning strategy with the robustness of large-scale pre-trained models, has set a new benchmark in the field of RGB Saliency Object Detection (SOD). Empirical evaluations across multiple datasets consistently underscored the superiority of our approach over extant methods. The distinctive blend of efficient parameter tuning and leveraging pre-existing architectures not only enhances the accuracy of saliency detection but also paves the way for future research, potentially guiding the development of more advanced and efficient SOD methodologies.

\bibliographystyle{splncs04}
\bibliography{ref}

\end{document}

%% file: files/benchmark.tex
\begin{tabular}{l|c|cc|cc|cc|cc|cc}
\toprule
\multicolumn{1}{c|}{\multirow{2}{*}{Method}} & \multicolumn{1}{l|}{\multirow{2}{*}{Year}} & \multicolumn{2}{c|}{PASCAL-S} & \multicolumn{2}{c|}{ECSSD} & \multicolumn{2}{c|}{HKU-IS} & \multicolumn{2}{c|}{DUTS-TE} & \multicolumn{2}{c}{DUT-OMRON} \\ \cline{3-12} 
\multicolumn{1}{c|}{}  & \multicolumn{1}{l|}{} & $F_{\beta} \uparrow$ & $\mathcal{M} \downarrow$ & $F_{\beta} \uparrow$ & $\mathcal{M} \downarrow$ & $F_{\beta} \uparrow$ & $\mathcal{M} \downarrow$ & $F_{\beta} \uparrow$ & $\mathcal{M} \downarrow$ & $F_{\beta} \uparrow$ & $\mathcal{M} \downarrow$ \\ \midrule
SCRN~\cite{scrn_wu}    & 2019  & .877  & .063  & .950  & .037  & .934  & .034  & .888  & .040  & .811  & .056  \\
GateNet~\cite{gatenet_zhao} & 2019  & .869  & .067  & .945  & .040  & .933  & .033  & .888  & .040  & .818  & .055  \\
ITSD~\cite{itsd_zhou}    & 2020  & .872  & .065  & .946  & .035  & .935  & .030  & .885  & .040  & .821  & .059  \\
LDF~\cite{ldf_wei}     & 2020  & .874  & .060  & .950  & .034  & .939  & .028  & .898                    & .034                         & .820                    & .052                         \\
CTDNet~\cite{ctd_zhao}                                       & 2021                                       & .878                    & .061                         & .950                    & .032                         & .941                    & .027                         & .897                    & .034                         & .826                    & .052                         \\
PFSNet~\cite{pfsnet_ma}                  & 2021                                       & .875                    & .063                         & .952                    & .031                         & .943                    & .026                         & .896                    & .036                         & .823                    & .055                         \\
EDN~\cite{edn_wu}                                          & 2022                                       & .880                    & \textbf{.062}                         & .951                    & .032                         & .941                    & \textbf{.026}                         & .895                    & .035                         & .828                    & 049                          \\
\midrule
SSOM                                         & -                                          & \textbf{.884}                    & .062                         & \textbf{.960}           & \textbf{.029}                & \textbf{.945}           & .027                & \textbf{.907}           & \textbf{.034}                & \textbf{.850}           & \textbf{.048}                \\ \bottomrule
\end{tabular}